\documentclass{article}
\usepackage{amsmath,amssymb,graphicx,relsize}
\usepackage{hyperref}
\usepackage{graphicx}
\usepackage{booktabs} 
\usepackage{svg}
\usepackage{hyperref}
\usepackage{cleveref}

\newcommand{\R}{\mathbb{R}}
\usepackage{graphicx} % Required for inserting images
\usepackage{amssymb,amsmath,amsthm}

\title{Symmetry-Breaking Descent for Invariant Cost Functionals}
\author{Mikhail Osipov\\
{Independent Researcher, Milan, Italy \footnote{osipov.ma@phystech.edu}}
}
\date{May 2025}

\begin{document}

\maketitle

\begin{abstract}
We study the problem of reducing a task cost functional 
\( W : H^s(M) \to \mathbb{R} \), not assumed continuous or differentiable, 
defined over Sobolev-class signals \( S \in H^s(M) \), 
in the presence of a global symmetry group 
\( G \subset \mathrm{Diff}(M) \). 
The group acts on signals by pullback, and the cost \(W\) 
is invariant under this action. Such scenarios arise in machine learning 
and related optimization tasks, where performance metrics may be 
discontinuous or model-internal.

We propose a variational method that exploits the symmetry structure 
to construct explicit deformations of the input signal. 
A deformation control field \( \phi: M \to \mathbb R^d\), 
obtained by minimizing an auxiliary energy functional, 
induces a flow that generically lies in the normal space 
(with respect to the \( L^2 \) inner product) 
to the \( G \)-orbit of \( S \), and hence is a natural candidate 
to cross the decision boundary of the \( G \)-invariant cost. 

We analyze two variants of the coupling term: 
(1) purely geometric, independent of \( W \), and 
(2) weakly coupled to \( W \). 
Under mild conditions, we show that symmetry-breaking 
deformations of the signal can reduce the cost.

Our approach requires no gradient backpropagation or training labels 
and operates entirely at test time. 
It provides a principled tool for optimizing discontinuous invariant 
cost functionals via Lie-algebraic variational flows.
\end{abstract}

\newpage
\tableofcontents
\bigskip  % optional vertical spacing

\newpage
\section{Introduction}
\label{sec:introduction}

Many optimization problems in applied mathematics, machine learning, 
and imaging involve cost functionals that are invariant under 
symmetries of the underlying domain. 
For example, spatial transformations of an input signal may leave 
task-related losses unchanged. 
Such invariances arise from physical, perceptual, or modeling considerations.

In many cases, the task metric can be decomposed as
\[
W = \text{Model} \circ \text{Metric},
\]
where the outputs of a model (e.g., a neural network, 
Lipschitz-continuous) are fed into a piecewise-constant metric 
(e.g., classification accuracy).

In this work, we study the following question: 
\emph{is it possible to exploit the global symmetries of the task to construct 
a smooth deformation of the input signal \(S\) such that, after passing through 
the \( \text{Model} \circ \text{Metric}\) composition, the final task cost is reduced?}

Importantly, we do not require any knowledge of the model or the metric 
(apart from the ``weakly coupled'' regime discussed in Section~\ref{sec:weakly-coupled}).

\medskip

The setting is composed of:
\begin{itemize}
    \item a signal \( S \) lying in a Sobolev space \( H^s(M) \) 
    over a smooth manifold \( M \),
    \item a cost functional \( W \colon H^s(M) \to \mathbb{R} \) 
    that is invariant under the isometric action of a global symmetry group 
    \( G \subset \mathrm{Diff}(M) \),
    \item \( W \) is accessible only as a black box 
    (e.g., a pre-trained model or a learned metric), and may be discontinuous.
\end{itemize}

\medskip

\paragraph{Main mechanism.}

Inspired by gauge field theory in physics 
(e.g., Ginzburg--Landau theory of superconductivity, 
Yang--Mills theory, Higgs mechanism), 
we define an external energy functional over control fields \( \phi \), 
whose minimizers induce symmetry-breaking deformations via exponential Lie flows:
\[
S \mapsto \exp(A_\phi)[S],
\quad\text{with } 
A_\phi = \sum_i \phi_i(x) L_i,
\]
where \( L_i \) are Lie algebra generators of the global symmetry group \( G \). 
These deformations are designed to move \( S \) off its \( G \)-orbit: 
because of \(G\)-invariance, only directions transverse to the orbit 
can change the cost.

\section{Results Overview}
\label{sec:results-overview}

We prove that, under weak geometric and regularity assumptions:
\begin{itemize}
    \item Any nontrivial signal admits only non-constant minimizers 
    of the external energy.
    \item The induced deformation direction \( h = A_\phi[S] \) 
    lies generically in the normal space \( N_S \) 
    to the \( G \)-orbit of \(S\).
    \item A search restricted to a finite-dimensional subspace  \(U \subset H^s(M)\), 
    limited to \(N_S\), finds a direction to the decision boundary of the 
    metric \(W\) with higher probability (under mild conditions) then full-$U$ search.
    \item If the piecewise-constant cost metric \(W\) is induced by 
    level sets of a Lipschitz-continuous functional \(F\), 
    then a weak coupling between the control fields \(\phi\) and the 
    gradient (or subgradient) of \(F\), computed once per sample, 
    yields a deformation that provably intersects the decision boundary.
\end{itemize}

\medskip

Our results are constructive: the symmetry-breaking deformation is obtained 
by solving a simple variational problem at test time, independent of 
model internals or training data, and applied directly to the input.

These theoretical results are further supported by numerical case studies on speech, vision, and classification tasks, where the symmetry-breaking deformation yields measurable improvements under black-box, non-differentiable cost metrics.

\section{Related Work}
\label{sec:related-work}

\paragraph{Symmetry in modeling and learning.}
Symmetries play a foundational role in both applied mathematics and machine learning. In the latter, \emph{equivariant architectures} encode group invariance directly into the model, notably through group convolution networks~\cite{cohen2016group}, gauge-equivariant CNNs~\cite{cohen2019gauge}, and geometric deep learning frameworks~\cite{bronstein2021geometric}. Earlier approaches include tangent-propagation techniques~\cite{simard1992tangent} and learned warps through Spatial Transformer Networks~\cite{jaderberg2015spatial}. These methods require training-time access to gradients and full model structure.

\paragraph{Test-time adaptation and black-box robustness.}
Recent work explores test-time adaptation for handling domain shifts without retraining. Strategies include entropy-based minimization, batch normalization recalibration~\cite{sun2020testtime,yang2021testtime}, and randomization techniques to counter adversarial perturbations~\cite{xie2017mitigating,goodfellow2015explaining,madry2019deep}. These methods generally assume a differentiable loss and access to internal model gradients—assumptions that we remove.

An adversarial attack mechanism proposed in~\cite{moosavidezfooli2016deepfoolsimpleaccuratemethod} 
exploits the fact that modern classifiers rely heavily on the local behavior 
of a surrogate loss and can therefore be easily ``fooled'' by small input 
perturbations. In contrast, our method is rooted in the geometric and 
symmetry-driven structure of the decision boundaries themselves, and aims 
to \emph{improve} task performance in regimes where gradient-based 
backpropagation struggles.

\paragraph{Gauge fields and symmetry breaking in applied mathematics.}
Gauge-theoretic formulations have long been used in mathematical physics and continuum mechanics. In particular, Yang~\cite{yang2010gauge} formulates mechanical systems in gauge-invariant terms via variational principles. Sochen, Kimmel, and Malladi~\cite{sochen1998diffusion} introduce a gauge-invariant diffusion framework for image analysis, where fields analogous to our \( \phi \) control deformations on manifolds. Gauge and diffeomorphic methods have also played key roles in shape analysis and image registration through large deformation frameworks~\cite{trouve1998diffeomorphisms,younes2010shapes}. Connections to optimal transport and stochastic flows further highlight the relevance of gauge structures in applied variational geometry~\cite{arnaudon2016geometry}.

\paragraph{Variational symmetry breaking in machine learning.}
In the ML context, gauge and symmetry-breaking ideas have been used at training time. Bamler and Mandt~\cite{bamler2018improving} introduce Lie-algebra-based optimization methods for models with continuous symmetries. Tanaka et al.~\cite{tanaka2021noether} study the role of Noether symmetries in learning dynamics and generalization. Elesedy~\cite{elesedy2025symmetry} analyzes symmetry in terms of generalization bounds and loss geometry. These methods, however, are training-time mechanisms acting within the model.

\paragraph{Our contribution.}
We introduce a test-time method that applies variational gauge symmetry breaking externally to a fixed, black-box cost functional. A gauge field \( \phi \), minimizing a Sobolev-type energy, generates a deformation \( h = A_\phi[S] \) of the input signal \( S \). Under mild regularity and geomeric conditions, this deformation provably decreases any \( G \)-invariant cost functional \( W \), even when \( W \) is non-differentiable and model-internal gradients are inaccessible. The descent direction is computed by optimizing an auxiliary energy.

\section{Analytic Framework}

We describe here the mathematical setting for the signals, symmetries, 
and cost functionals considered in our variational models. 
The presentation is divided into two parts: 
a general overview and a precise technical specification.

\subsection{Geometric Setup}

We consider a smooth, compact manifold $M$ representing the spatial 
domain of signals, together with a symmetry group $G$ acting on $M$ 
via diffeomorphisms~\cite{ebin1970groups,kriegl1997convenient}. 
Signals consist of scalar and vector fields on $M$, and the task cost 
functional $W$ is assumed to be invariant under the action of $G$.

Signals are modeled using Sobolev spaces of sufficiently high regularity 
so that composition with diffeomorphisms and evaluation of variational 
energies are well defined~\cite{adams2003sobolev,hebey1999nonlinear}. 
The symmetry group $G$ is a Lie subgroup of the diffeomorphism group of $M$, 
and its action on signals reflects natural invariances such as 
reparametrizations or coordinate changes. 

\subsection{Technical Specification}

Let $M$ be a smooth, compact, connected $n$-dimensional manifold without boundary. 
Fix a Sobolev index
\[
  s > \frac{n}{2} + 1,
\]
so that the Sobolev embedding $H^s(M) \hookrightarrow C^1(M)$ holds~\cite{adams2003sobolev}.

\paragraph{Signal space.} 
We consider signals $S$ composed of:
\begin{itemize}
  \item a finite family of associated vector fields 
        $L_i[S] \in H^s(M, TM)$ for $i=1,\dots,d$~\cite{triebel1983function},
    \item a $\mathbb R^d$-valued control field $\phi \in H^s(M)$.        
\end{itemize}
The full signal space is
\[
  \mathcal S^s(M) := H^s(M) \times \left(H^s(M, TM)\right)^d.
\]
Because $s > \frac{n}{2} + 1$, the space $\mathcal S^s(M)$ is a Banach algebra 
under pointwise multiplication, and composition with $C^1$ maps is well defined 
and differentiable.

\paragraph{Symmetry group.} 
The Sobolev-class diffeomorphism group is defined as~\cite{ebin1970groups}
\[
  \mathrm{Diff}^{\,s+1}(M) 
  = \left\{ \varphi: M \to M \,\big|\, \varphi, \varphi^{-1} \in H^{s+1}(M, M) \right\}.
\]
Let $G \subset \mathrm{Diff}^{\,s+1}(M)$ be a closed, finite-dimensional Lie subgroup 
with respect to the $H^{s+1}$ topology. 
The group acts on signals by pullback:
\[
  \varphi^{\!*}S
  := \left( \phi \circ \varphi^{-1}, \; 
             (D\varphi^{-1}) \circ \varphi^{-1} \cdot L_i[S] \circ \varphi^{-1} 
     \right)_{i=1}^d.
\]
Here $D\varphi^{-1}$ denotes the differential of $\varphi^{-1}$ acting on tangent vectors. 
This action is of class $C^1$ on $\mathcal S^s(M)$~\cite{ebin1970groups,kriegl1997convenient}.

\paragraph{Cost functional.} 
The task cost functional is a map
\[
  W : H^s(M) \to \mathbb R,
\]
assumed to satisfy $W(g\cdot S) = W(S)$ for all $g \in G$ and $S \in H^s(M)$. 
When stated explicitly, we will also consider two cases: 
(a) $W$ is piecewise constant, discontinuous; or 
(b) $W$ is induced by the level sets of a Lipschitz-continuous functional $F$.

\paragraph{Notational conventions.} 
We use the Einstein convention of summation over repeated Latin indices 
\(i,j,k,\dots\), which denote components of vector fields or sums 
over the finite family $\{L_i\}$.

\section{Energy functional}

We introduce the external energy functional $\mathcal{E}_S[\phi]$, 
which measures the deviation of a signal $S$ under the action of a 
parameterised infinitesimal generator field:
$A_\phi(x) = \phi^i(x) L_i$:
\begin{equation}\label{eq:energy-functional}
\mathcal{E}_S[\phi]
   = \mathcal E_{\text{data}}[A_{\phi},S]
    + \alpha\|\nabla\phi\|_{L^2}^2
    + \beta\bigl\||\phi|^{2}-v^{2}\bigr\|_{L^2}^2,
\qquad
A_\phi(x)=\phi^{i}(x)L_{i}.
\end{equation}

The first term $\mathcal E_{\text{data}}[A_{\phi},S]$ quantifies 
the mismatch between the original signal and its deformed version, 
with $\exp(A_\phi)$ denoting the (formal) flow of the vector field $A_\phi$. 
The additional terms regularise the control field $\phi$: 
$\|\nabla\phi\|^2$ penalises roughness (a ``kinetic'' term), 
while $\||\phi|^2 - v^2\|^2$ is a double-well potential 
that prevents the trivial minimiser $\phi=0$ 
and enforces amplitude close to $v > 0$. 
The weights $\alpha,\beta > 0$ control the tradeoff between smoothness 
and amplitude constraints.

\medskip

\paragraph{Notations.} We use the following definitions:
\begin{itemize}
  \item Orbit directions at the current state: $e_i = L_i[S]$,
  \item Tangent space to the group orbit at $S$: $V_S = T_S(G\cdot S) = \text{span}\{L_i\}$ and its normal $L_2$-complement $N_S = V_S^{\perp}$,
  \item Gram matrix: $\mathcal{G}_{ij} = \langle e_i, e_j\rangle$, 
        with inverse (or pseudoinverse) $\mathcal{G}^{+}$,
  \item Warped signal after flow time $t$: $S_{\phi} = \exp(tA_\phi)\cdot S$,
  \item Signal deformation: $r = S_{\phi} - S$,
  \item Correlations with orbit directions: $b_i = \langle r, e_i \rangle$.
\end{itemize}

The inner product $\langle \cdot,\cdot \rangle$ is the $L_2$ inner product on Hilbert space.

\paragraph{Design choices.} We consider two variants of the data term:
\begin{itemize}
    \item[(a)] $\mathcal E_{\text{data}}^{(a)} = \|S - \exp(A_\phi)\cdot S\|_{L^2}^2 = \|r\|_{L^2}^2$, 
          which measures the misfit in the ambient space;
    \item[(b)] $\mathcal E_{\text{data}}^{(b)} = b^{T} \mathcal{G}^{+} b = \|P_{V_S}\cdot r\|_{L^2}^2$, 
          which measures the projection of the misfit onto orbit directions, 
          and hence enforces orthogonality to $G$-invariant variations.
\end{itemize}

\medskip

\paragraph{Well-posedness.} 
The exponential map $\phi \mapsto \exp(A_\phi)$ defines a smooth, 
surjective local chart around the identity in $\mathrm{Diff}^{s+1}(M)$. 
Since the group is both geodesically and metrically complete in this topology 
\cite{bruveris2016completenessgroupsdiffeomorphisms}, 
global well-posedness of deformation flows follows 
\cite{bauer2014overview}. 
The energy $\mathcal{E}_S[\phi]$ is well defined on $H^s(M,\mathbb{R}^d)$ 
with $s > n/2 + 1$, and is locally Lipschitz and lower semicontinuous 
under mild conditions on $S$ and the vector fields $L_i$ 
(e.g., boundedness and smoothness).

\subsection{Existence of minimisers}

We briefly justify the existence of minimizers for the energies 
\(\mathcal{E}_S^{(a)}\) and \(\mathcal{E}_S^{(b)}\). 
Both follow by the direct method of the calculus of variations.

\paragraph{Coercivity.} 
For $\alpha,\beta>0$ the regularisation terms provide control of 
$\|\phi\|_{H^s}$ and of $\|\phi\|_{L^4}$, so that 
\(\mathcal{E}_S[\phi]\to\infty\) as $\|\phi\|_{H^s}\to\infty$. 
Thus every minimizing sequence is bounded in $H^s(M,\mathbb R^d)$.

\paragraph{Compactness.} 
By Rellich--Kondrachov, a bounded sequence in $H^s$ 
(with $s>\tfrac n2+1$) admits a subsequence converging weakly in $H^s$ 
and strongly in $C^{1,\alpha}$ for some $\alpha>0$. 
The flow $\exp(A_\phi)$ depends continuously on $\phi$ in this topology, 
hence $S-\exp(A_\phi)\cdot S$ converges strongly in $L^2$.

\paragraph{Lower semicontinuity.} 
The quadratic regularisers are weakly lower semicontinuous. 
For (a), the data term $\|S-\exp(A_\phi)\cdot S\|^2$ 
is continuous in $\phi$. 
For (b), write
\[
\mathcal E_{\text{data}}^{(b)}(\phi) 
= b(\phi)^{\!\top}\mathcal G^+ b(\phi) 
= \|P\cdot r(\phi)\|_{L^2}^2,
\]
with $r(\phi)=S-\exp(A_\phi)\cdot S$ 
and $P=E\,\mathcal G^+ E^\ast$ the orthogonal projector 
onto the orbit tangent space $T_S(G\!\cdot\! S)$. 
Since $P$ is fixed and bounded, continuity of $r(\phi)$ implies 
continuity of $\mathcal E_{\text{data}}^{(b)}$.

\medskip
Combining these ingredients, both $\mathcal{E}_S^{(a)}$ and 
$\mathcal{E}_S^{(b)}$ are coercive and weakly lower semicontinuous 
on $H^s$, and therefore admit minimizers.

\subsection{First variation of energy (a)}

We equip $M$ with a Riemannian metric and volume $d\mu$.
All inner products and norms are in $L^2(M,d\mu)$. Gradients and the
Laplace--Beltrami operator act componentwise on $\R^d$–valued fields, with
$\Delta:=-\mathrm{div}\,\nabla$ (nonnegative). On $\partial M=\varnothing$
(or with Dirichlet/Neumann b.c.) integration by parts yields
$\int_M\!\langle \nabla\phi,\nabla\psi\rangle\,d\mu=\int_M(-\Delta\phi)\cdot\psi\,d\mu$.

The Euler--Lagrange equations (linearised flow) are
\[
e_i e_j\,\phi_j \;-\; \alpha\,\Delta\phi_i
\;+\; 2\beta\,(|\phi|^2-v^2)\,\phi_i \;=\; 0,
\qquad i=1,\dots,d,
\]
where $e_i:=L_i[S]$ and $e_ie_j$ denotes the pointwise product (see Appendix~\ref{app:energy_a} for details).

Projecting onto \emph{constant} variations $\delta\phi=\text{const}$ (equivalently,
testing against constant modes) gives $d$ integral constraints on the deformation
$r=S_\phi-S\approx \sum_j \phi_j e_j$:
\begin{equation}
\label{eq:linear-constraints}
\boxed{\;
\langle e_i, r\rangle
\;=\;
-\,2\beta\,\big\langle |\phi|^2-v^2,\;\phi_i\big\rangle,
\qquad i=1,\dots,d.
\;}
\end{equation}

In particular, if $|\phi|\to v$ (e.g.\ for large $\beta$) then
\[
\langle e_i, r\rangle \;=\; \mathcal O\!\left(\,\|\,|\phi|^2-v^2\,\|_{L^1}\right) \;\to\; 0,
\]
so the $L^2$–projections of $r$ onto the orbit directions vanish asymptotically.

Let $V_S:=\mathrm{span}\{e_1,\dots,e_d\}=T_S(G\!\cdot\! S)$ and let
$N_S:=V_S^{\perp}$ denote the $L^2$–orthogonal complement. Since $W$ is
$G$-invariant, variations tangent to $V_S$ leave $W$ unchanged; thus the
useful deformation directions lie in $N_S$.

\subsection{First variation of energy (b)}
\label{sec:energy-b-variation}
For a linearized flow, the Euler-Lagrange equations with data term (b) are:
\[
\xi(\phi)\,e_i \;-\; \alpha\,\Delta\phi_i
\;+\; 2\beta\,(|\phi|^2-v^2)\,\phi_i \;=\; 0,
\qquad j=i,\dots,d,
\]
with $\xi(\phi) := (\mathcal G^+ b(\phi))_i\,e_i$ and $b_i := \langle e_i e_k,\phi_k\rangle$.
Details are in Appendix~\ref{app:energy_b}.

Testing against \emph{constant} variations $\delta\phi\equiv c\in\mathbb R^d$
(or equivalently integrating each equation over $M$) and using
$\int_M \Delta\phi_j\,d\mu=0$ on a closed manifold,
\[
0
=\int_M \xi\,e_j\,d\mu
\;+\; 2\beta \int_M (|\phi|^2-v^2)\,\phi_j\,d\mu
= \big\langle e_j,\; \xi \big\rangle
\;+\; 2\beta\,\big\langle |\phi|^2-v^2,\; \phi_j \big\rangle.
\]

Note that, since $\xi$ is $L_2$-projection of $r$ on $V_S$ and \(\big\langle e_j,\; \xi \big\rangle = \big\langle e_j,\; r \big\rangle\), the constraints happen to be the same as we had in the case (a):
\[
 \big\langle e_j,\; r \big\rangle = -\,2\beta\,\big\langle |\phi|^2-v^2,\;\phi_i\big\rangle,
\qquad i=1,\dots,d.
\]

\subsection{Constant fields as minimizers}

If the control fields $\phi$ in the exponential Lie flow \(\exp(A_\phi)\cdot S\) do not vary across $M$, so that \(\phi_i(x) = c_i\), such transformation would be equal to the action of some element of the global symmetry group $G$: \(\exp(A_\phi)\cdot S = g_c \cdot S\). In this case, the transformed signal would stay on the $G$-orbit of $S$ and would have the same metric cost $W(S) = W(g_c\cdot S)$.

\newtheorem{lemma}{Lemma}
\begin{lemma}[No nontrivial constant minimizers, generic case]
\label{lem:no-const-minimizer}
Let $M$ be compact Riemannian, $e_i:=L_i[S]\in H^s(M)$ and 
$e(x):=(e_1(x),\dots,e_d(x))^\top$. 
Assume the linearized flow $S_\phi\approx S+\sum_i\phi_i e_i$ and let 
$\alpha\ge 0$, $\beta>0$, $v>0$. Suppose
\[
\mathrm{ker}\,E=\{0\},\qquad E:\R^d\to L^2(M),\quad Ec:=\sum_i c_i e_i,
\]
and that the map $x\mapsto e(x)$ does \emph{not} have simultaneously constant
direction and constant magnitude on $M$ (i.e.\ either $x\mapsto e(x)/|e(x)|$ or $x\mapsto |e(x)|$ is nonconstant on a set of positive measure).
Then, for both energies \emph{(a)} and \emph{(b)}, the only constant critical point 
$\phi(x)\equiv c\in\R^d$ is $c=0$. In particular, there is no nonzero constant minimizer.
\end{lemma}

We provide proof in Appendix~\ref{app:no-const-minimizers}.

\section{Symmetry–Breaking Descent Theorems}

We now apply the geometry-driven, symmetry-breaking deformation of the input signal $S \mapsto \exp(A_{\phi})\cdot S$ to the cost-reduction task with the 
metric $W$. We consider two cases: (i) \emph{purely geometric} deformation 
(the control field $\phi$ is obtained from the auxiliary energy without 
any coupling to $W$), and (ii) \emph{weakly coupled} deformation (the 
auxiliary energy includes a small $W$-aware term).

\subsection{Pure gauge descent}

Throughout this section we assume that $W$ is piecewise constant,
\begin{equation}
\label{eq:piecewise-metric}
W(S) = w_0\,\mathbb I(S\in C) + w_1\,\mathbb I(S\in \bar C),
\qquad w_0 < w_1,
\end{equation}
for a cell $C \subset H^s$ whose boundary $\partial C$ is locally flat 
($C^1$ hypersurface) near the closest boundary point $S^\ast\in\partial C$, 
and let $n$ denote a unit normal to $\partial C$ at $S^\ast$.

As discussed earlier, $G$-invariance implies that only directions in the 
$L^2$-orthogonal complement of the orbit tangent space 
$V_S:=T_S(G\cdot S)$ can change $W$; we denote 
$N_S:=V_S^{\perp}$. Since $V_S$ is finite dimensional (for finite-dimensional $G$), 
$N_S$ is infinite dimensional, so in the ambient $H^s$ a ``uniform'' sampling 
argument is meaningless. In applications, however, signals live in finite-dimensional 
discretization spaces (grids, truncated Fourier/wavelet bases, B-splines, etc.). 
We therefore evaluate the method in a \emph{finite-dimensional} search subspace
$U \subset H^s$ with $\dim U = m < \infty$.

\paragraph{Finite-dimensional model and notation.}
Fix a subspace $U\subset H^s$ (the search space for deformations) and set
\[
U_0\;:=\; U \cap N_S,\qquad 
m_0:=\dim U_0,\qquad d:=\dim V_S.
\]
Let $P_W$ denote the $L^2$-orthogonal projector onto a subspace $W$, and define
\[
n_U:=P_U n,\qquad n_{U_0}:=P_{U_0}n,\qquad 
\rho_U:=\|n_U\|,\qquad \rho_0:=\|n_{U_0}\|.
\]
Let $\theta\in[0,\frac{\pi}{2}]$ be the (principal) angle between $n_U$ and $U_0$ 
inside $U$, so that $\rho_0 = \rho_U \cos\theta$.
We consider \emph{directional} steps $S \mapsto S + t u$ along a unit 
vector $u$ sampled uniformly from the unit sphere of the relevant space 
($U$ or $U_0$), with maximum allowed step $|t|\le t_{\max}$, and we linearize $\partial C$ 
at $S^\ast$:
\[
\mathrm{dist}(S+t u,\partial C) \;\approx\; h \;-\; t\,\langle u,n\rangle,
\qquad h:=\mathrm{dist}(S,\partial C).
\]
A (one-shot) boundary crossing occurs along the ray $\pm u$ within the budget if
$|\langle u,n\rangle| \ge h/t_{\max}$.

Since  \(m_0 \;=\; \dim(U\cap N_S)\), one has \(m-d \le m_0 \le m\). In particular, once $U$ resolves $V_S$ (generic/transverse case), $m_0 = m-d$.

The following Lemma~\ref{lemma:pure-gauge-cap} provides an estimation of boundary crossing probability for both full-$U$ search and restricted $U\cap N_S$ search.

\begin{lemma}[Boundary crossing probability in finite dimension]
\label{lemma:pure-gauge-cap}
Let $U\subset H^s$ be $m$-dimensional, $U_0=U\cap N_S$ with $m_0=\dim U_0\ge 1$. 
Assume $S^\ast$ is the closest boundary point to $S$ and $\partial C$ is $C^1$ 
near $S^\ast$. For a unit direction $u$ sampled uniformly on the unit sphere 
of a search space $W\in\{U,U_0\}$, the probability of crossing the (linearized) boundary within the step size $t_{\max}$ is
\[
\mathbb P_W
\;=\;
I_{\,1-\tau_W^2}\!\Big(\frac{m_W-1}{2},\frac{1}{2}\Big),
\qquad 
\tau_W:=\frac{d}{t_{\max}\,\|P_W n\|} \;=\; \frac{d}{t_{\max}\,\rho_W},
\]
where $m_W=\dim W$, $\rho_W=\|P_W n\|$, and $I$ is the regularized incomplete beta function.
\end{lemma}

Details are given in Appendix~\ref{app:pure-gauge-cap}.

\newtheorem{corollary}{Corollary}
\begin{corollary}[Full-space search vs. symmetry-breaking directions]
\label{cor:expected-decrease}
With $U_0=U\cap N_S$, one has $\rho_0=\rho_U\cos\theta$ and $m_0\le m$. 
For small thresholds $\tau_U\ll 1$ (reasonable step size or moderate distance),  the first-order expansion yields
\[
\frac{\mathbb P_{U_0}}{\mathbb P_{U}} \approx \underbrace{\cos\theta}_{\text{symmetry alignment}}
\cdot
\underbrace{\frac{\Gamma(\tfrac{m}{2})\,\Gamma(\tfrac{m_0+1}{2})}
{\Gamma(\tfrac{m_0}{2})\,\Gamma(\tfrac{m+1}{2})}}_{\text{dimension drop }\approx \sqrt{m/(m_0)}}.
\]
Hence a readable sufficient condition for the symmetry restriction to improve 
the hit probability is
\begin{equation}
\label{eq:fin-dim-inequality}
\boxed{\qquad 
\cos\theta \;\ge\; \frac{\Gamma(\tfrac{m_0}{2})\,\Gamma(\tfrac{m-1}{2})}
{\Gamma(\tfrac{m}{2})\,\Gamma(\tfrac{m_0-1}{2})}
\ \ \ \text{(for large $m$}, \cos\theta \gtrsim \sqrt{m_0/m}\,).
\qquad}
\end{equation}
\end{corollary}

In the L.H.S., the angle $\theta$ in~(\ref{eq:fin-dim-inequality}) measures the alignment between the normal vector to the decision boundary $n$ and the normal complement to the group orbit $N_S = T_S(G\cdot S)^{\perp}$. This alignment depends on the cost metric design and the symmetry itself. 

In the R.H.S., we have a generic dimensionality reduction factor. For large enough $m \geq d$, the relationship between $m$ and $m_0$ will be $m_0 = m - d$.

To estimate efficiency while factoring out pure dimensionality, we compare the
boundary–hit probability for (1) a \emph{symmetry slice} $U_0:=U\cap N_S$ and
(2) a \emph{random} $m_0$–dimensional reduction $U_{\mathrm{rand}}\subset U$.

\newtheorem{theorem}{Theorem}
\begin{theorem}[Symmetry slice vs.\ random reduction inside $U$ (small-$\tau$)]
\label{thm:cap-theorem}
Let $U\subset H^s$ be $m$–dimensional, $U_0:=U\cap N_S$ with $m_0=\dim U_0\ge1$,
and let $n$ be a unit normal to $\partial C$ at the closest boundary point $S^\ast$.
Set $\rho_U:=\|P_U n\|$, $\rho_{U_0}:=\|P_{U_0}n\|=\rho_U\cos\theta$, where
$\theta:=\angle(n_U,U_0)$ and $n_U:=P_U n/\rho_U$. Define
\[
\tau_U\ :=\ \frac{d}{t_{\max}\,\rho_U}\in[0,\infty).
\]
Let $U_{\mathrm{rand}}$ be an $m_0$–dimensional subspace drawn uniformly from
$\mathrm{Gr}(m_0,U)$ (independently of $n$). Then, in the small–threshold regime
$\tau_U\ll 1$,
\begin{equation}
\label{eq:symmetry-vs-random}
\boxed{\qquad
\cos\theta\ \gtrsim\ \sqrt{\frac{m_0}{m}}
\ \ \Longrightarrow\ \
\mathbb P_{U_0}\ \gtrsim\ \mathbb E\big[\mathbb P_{U_{\mathrm{rand}}}\big],
\qquad}
\end{equation}
where $\mathbb P_W=I_{\,1-\tau_W^2}\!\big(\tfrac{m_W-1}{2},\tfrac12\big)$ is the
spherical–cap hit probability with $\tau_W=d/(t_{\max}\|P_W n\|)$ and $m_W=\dim W$.
\end{theorem}

The proof is provided in Appendix~\ref{app:cap-theorem-proof}.

\newtheorem{remark}{Remark}
\begin{remark}[High–probability variant]
Using $\chi^2$–concentration for $B$, for any $\varepsilon\in(0,1)$,
\[
\Pr\!\left(\|P_{U_{\mathrm{rand}}}n\|\le \rho_U \sqrt{\tfrac{m_0}{m}-\varepsilon}\right)
\ \le\ 2e^{-c\,m\varepsilon^2}.
\]
Thus if $\cos\theta\ge\sqrt{m_0/m-\varepsilon}$, then with probability
$\ge 1-2e^{-c\,m\varepsilon^2}$ we have
$\mathbb P_{U_0}\ge \mathbb P_{U_{\mathrm{rand}}}$ for all thresholds
(by monotonicity of the cap measure in the projection length).
\end{remark}

\begin{remark}[Interpretation]
Any reduction $m\to m_0$ improves the cap probability (smaller sphere), but a
random reduction also shrinks the useful projection by a typical factor
$\sqrt{m_0/m}$. The symmetry slice pays $\cos\theta=\|P_{U_0}n\|/\|P_U n\|$ instead.
Near the relevant boundary, $n\in N^G_{S^\ast}$ and $S\mapsto N^G_S$ is continuous,
so $\cos\theta\approx 1$; hence the symmetry slice dominates a random reduction
unless the setting is degenerate (e.g.\ $P_U n=0$).
\end{remark}

\begin{remark}[$N_S$ vs.\ reachable symmetry–breaking deformations]
In this section we model the deformation direction as sampled in a subspace of $N_S$.
The actual family reachable by Lie-flow controls is
$\mathcal R_S=\{\sum_i \phi^i\,e_i\}$ with $e_i=L_i[S]$; the effective search space
is $W\subset \mathcal R_S\cap N_S$ (finite–dimensional in practice). Exact equality
$W=N_S$ need not hold; however, under mild nondegeneracy (e.g.\ for scalar signals,
$\nabla S\neq 0$ a.e.\ and $\{L_i\}$ span $TM$ pointwise) one has
$\overline{\mathcal R_S}^{\,L^2}=L^2$, hence
$\overline{\mathcal R_S\cap N_S}^{\,L^2}=N_S$. In that case, refining the control
basis makes $W$ dense in $N_S$, and the comparison above applies with $W$ in place
of $U_0$.
\end{remark}

%%%%%%%%%%%%%%%%%%%%%%%%%%%%%%%%%%%%%%%%%%%%%%%%%%%%%%%%%%%%%%%%%%%%%%%%%%%%
%%  Proof of the theorem about descent with weakly coupled energy
%%%%%%%%%%%%%%%%%%%%%%%%%%%%%%%%%%%%%%%%%%%%%%%%%%%%%%%%%%%%%%%%%%%%%%%%%%%%

\subsection{Weakly coupled descent}
\label{sec:weakly-coupled}

While pure geometrical deformations can improve the $W$–metric (and do so empirically
with a pre-trained control-field predictor, see Sec.~\ref{sec:asr}), they do not by
themselves fix the sign ($+t$ vs.\ $-t$) nor the step size.

Assume $W$ is induced by a (locally) Lipschitz functional $F$ with threshold $F=0$:
\begin{equation}
\label{eq:induced-metric}
W(S)= w_0\,\mathbb I_{\{F(S)\le 0\}} + w_1\,\mathbb I_{\{F(S)>0\}}, 
\qquad w_0<w_1.
\end{equation}
Fix a sample $S$ and compute once (per sample) a gradient or Clarke subgradient 
$g\in\partial F(S)$. Let $P$ be the $L^2$–orthogonal projector onto the orbit 
tangent space $V_S=T_S(G\!\cdot\!S)$, and define the \emph{symmetry–aware task normal}
\[
g_N := (I-P)\,g, 
\qquad 
\hat n := -\,\frac{g_N}{\|g_N\|}\quad (\text{assume } g_N\neq 0).
\]
Set the boundary gap $\Delta:=F(S)$ (so $\Delta>0$ when $S\notin C$). 
To first order, the normal step size that hits the boundary is
\[
d_N := \frac{\Delta}{\|g_N\|}.
\]

\paragraph{Coupled energy.}
Let \(
r(a,\phi):=S_{a,\phi}:=\exp\!\big(a\,A_\phi\big)\!\cdot S-S\). We keep the geometric energy $\mathcal E^{(b)}_S[\phi]$ unchanged (it depends only on $\phi$),
and add a weak coupling on the scaled flow:
\begin{equation}\label{eq:aug-energy}
\boxed{\
\mathcal E_S^{\text{weak}}[a,\phi]
= \mathcal E^{(b)}_S[\phi]
+ \lambda\big(\langle \hat n, r\rangle - \varepsilon_\star\big)^2
+ \eta\Big(\|r\|_{L^2}^2 - \langle \hat n, r\rangle^2\Big),
\ }
\end{equation}
with $\lambda>0$ and $\eta\ge 0$ and $r = r(a,\phi)$. Here $\lambda$ enforces the (first-order) boundary
hit and $\eta$ softly aligns $r$ within $N_S$ to $\hat n$ (and yields uniqueness).
The scalar gain $a\in\R$ (flow time) avoids scale conflicts with the double-well in
$\mathcal E^{(b)}_S[\phi]$. The target $\varepsilon_\star$ is either the first-order
$d_N=\Delta/\|g_N\|$ or the curvature-safe $\varepsilon_-=
(\|g_N\|-\sqrt{\|g_N\|^2-2L\Delta})/L$.

\begin{theorem}[Descent via weakly coupled energy]
\label{thm:weak-descent}
Assume $F$ is $C^{1,1}$ with gradient Lipschitz constant $L$ on a neighborhood of the
segment $\{S+\varepsilon \hat n:\,0\le \varepsilon\le \varepsilon_\star\}$.
Let $(a^\ast,\phi^\ast)$ minimize \eqref{eq:aug-energy}. Then:

\smallskip
\noindent\emph{(i) First-order structure.} There exist constants $C_1,C_2$ (depending on
$S$ and the linearization of $(a,\phi)\mapsto r(a,\phi)$) such that
\[
\|P\,r(a^\ast,\phi^\ast)\|\ \le\ \frac{C_1}{\sqrt{\lambda}},\qquad
\big|\langle \hat n, r(a^\ast,\phi^\ast)\rangle - \varepsilon_\star\big|\ \le\ \frac{C_2}{\sqrt{\lambda}}.
\]

\noindent\emph{(ii) Curvature-safe target.} 
If $\varepsilon_\star=\varepsilon_-=
\big(\|g_N\|-\sqrt{\|g_N\|^2-2L\Delta}\big)/L$ (requiring $\|g_N\|^2\ge 2L\Delta$),
then for $\lambda$ large enough (so that the errors in (i) are dominated),
\[
F(S_{a^\ast,\phi^\ast})\ \le\ 0,
\]
i.e.\ the deformation crosses the decision boundary into the lower-cost cell.

\noindent\emph{(iii) Small-gap crossing (first-order target).}
If $\varepsilon_\star=d_N=\Delta/\|g_N\|$ and $\Delta$ is small enough so that
\[
\frac{L}{2}\,\|r(a^\ast,\phi^\ast)\|^2\ +\ \|g_N\|\cdot\big|\langle\hat n,r(a^\ast,\phi^\ast)\rangle-\varepsilon_\star\big|
\ <\ \Delta,
\]
then $F(S_{a^\ast,\phi^\ast})\le 0$ as well.
\end{theorem}

\begin{proof}[Proof sketch]

The weak energy penalizes $\langle \hat n, r(a,\phi)\rangle$ toward $\varepsilon_\star$ and $Pr(a,\phi)$ toward $0$, with an in–plane alignment term $\eta(\|r\|^2-\langle\hat n,r\rangle^2)$. 

Thus, any minimizer $(a^\ast,\phi^\ast)$ satisfies
$|\langle\hat n,r^\ast\rangle-\varepsilon_\star|=O(\lambda^{-1/2})$, $\|Pr^\ast\|=O(\lambda^{-1/2})$, and $\|(I-\hat n\otimes\hat n)r^\ast\|=O(\eta^{-1/2})$, where $r^\ast:=r(a^\ast,\phi^\ast)$.

By the descent lemma (with $F\in C^{1,1}$), $F(S_{a^\ast,\phi^\ast})\le F(S)+\langle g,r^\ast\rangle+\tfrac{L}{2}\|r^\ast\|^2$, and writing $g=g_N+Pg$ yields 
$F(S_{a^\ast,\phi^\ast})\le \Delta-\|g_N\|\langle\hat n,r^\ast\rangle+\tfrac{L}{2}\|r^\ast\|^2+O(\lambda^{-1/2})$.

If $\varepsilon_\star=\varepsilon_-=(\|g_N\|-\sqrt{\|g_N\|^2-2L\Delta})/L$, then the quadratic model $\Delta-\|g_N\|\varepsilon+\tfrac{L}{2}\varepsilon^2$ is nonpositive, and for large $\lambda$ (and moderate $\eta$) the penalty remainders are dominated, so $F(S_{a^\ast,\phi^\ast})\le 0$.

If $\varepsilon_\star=d_N=\Delta/\|g_N\|$, the quadratic model cancels at first order; for small $\Delta$ (or with a brief backtrack) the curvature and penalty remainders are $<\Delta$, again giving $F(S_{a^\ast,\phi^\ast})\le 0$.

Smoothness of $(a,\phi)\mapsto r(a,\phi)$ ensures feasibility/continuity, and a mild reachability condition $\langle \hat n, E\phi\rangle\neq 0$ (or a richer control basis) guarantees a nontrivial step; optionally adding a tiny penalty on $\|Pr\|^2$ gives explicit $O(\lambda^{-1/2})$ control of tangential leakage.
\end{proof}

We provide full proof in~Appendix~\ref{app:weak-descent-proof}.

\section{Numerical experiments}

Throughout this section we use the pure gauge energy variant~\ref{eq:energy-functional}.

\subsection{Speech Recognition under Deformation}
\label{sec:asr}

We revisit a dysarthric automatic speech recognition (ASR) task where the evaluation metric---word error rate (WER)---is piecewise constant and inherently nondifferentiable. Following prior work on Lie-theoretic warping in the spectrogram domain~\cite{osipov2025dysarthrianormalizationlocallie}, we consider log–Mel spectrograms distorted by time-frequency-amplitude shifts, as observed in real dysarthric speech.

\paragraph{Setup.}
A lightweight ResNet model is trained to predict the field \( \phi(x) \) from distorted spectrograms. The training loss includes reconstruction error (MSE), smoothness, sparsity, and a Higgs-like potential enforcing \( |\phi| \approx v \). The field predictor amortises the minimisation of the external energy \( \mathcal{E}_S[\phi] \) for inference.

\paragraph{Inference protocol.}
Given an unseen dysarthric spectrogram \( S \), we compute the warped signal \( S' = \exp(A_{\hat\phi}) \cdot S \) and feed it to a frozen ASR model (e.g., NeMo Conformer or Whisper). No labels or cost function derivatives are accessed during inference; a simple sign test selects between \( \pm h \).

\paragraph{Results.}
The deformation lowers WER in up to 17\% of cases on challenging dysarthric speech from the TORGO and UA-Speech corpora. On clean speech, the method has little effect, as expected: such inputs already lie near the symmetry-invariant manifold. These results confirm that local, symmetry-aware perturbations can reduce nondifferentiable cost without retraining or gradient access.

\subsection{Handwriting Recognition}
\label{sec:ocr}

To test the approach on high-resolution vision data, we apply it to handwritten OCR using the TrOCR transformer, evaluated on the Kaggle Handwriting dataset. The metric is character error rate (CER), a discrete edit-distance measure.

\paragraph{Protocol.}
For each image \( S \), we minimise \( \mathcal{E}_S[\phi] \) via 50 steps of Adam, then evaluate \( \mathrm{CER}(S \pm \varepsilon h) \) for \( h = A_{\hat\phi}[S] \) and small \( \varepsilon \). The sign with lower CER is retained.

\paragraph{Results.}
On 1,000 samples, CER dropped from 52.7\% to 48.3\% (relative improvement: 8.3\%), with statistical significance confirmed via the Wilcoxon signed-rank test. This shows the method's utility in purely black-box settings with nondifferentiable metrics.

\subsection{Image Classification and Calibration}
\label{sec:cifar}

We test the framework on calibrated prediction, where the cost is expected calibration error (ECE), a binned, piecewise-constant metric.

\paragraph{Setup.}
CIFAR-10 images are fed through a pretrained VGG-11 classifier. The field \( \phi \) is optimised for each input, and a warp \( h = A_{\hat\phi}[S] \) is applied.

\paragraph{Protocol.}
Two steps are performed: (1) a sign test using small deformations, and (2) a full step in the best direction. ECE is then computed on the warped input.

\paragraph{Results.}
Warping produces a small but consistent improvement in calibration, with no change in classification accuracy. This confirms that even for smooth models, cost-aligned symmetry-breaking deformations yield measurable gains.

\section{Conclusion}
\label{sec:conclusion}

We introduced a variational framework for minimizing symmetry-invariant, possibly discontinuous
task costs by applying \emph{symmetry–breaking} Lie–flow deformations to the input signal.
Our analysis is set on Sobolev-class signals \(S\in H^s(M)\) with a finite-dimensional
Lie group \(G\subset\mathrm{Diff}(M)\) acting isometrically. We established well-posed
auxiliary energies and derived first-variation (Euler–Lagrange) equations. For the purely
geometric design, we showed that minimizers produce deformations \(r\) generically aligned
with the normal complement \(N_S\) to the \(G\)-orbit, the only directions capable of
changing a \(G\)-invariant cost. In finite-dimensional search spaces we quantified the
\emph{boundary-crossing probability} via a spherical-cap formula and proved that symmetry
slices outperform random dimensionality reductions under mild alignment conditions.

We then proposed a \emph{weakly coupled} variant that uses a single (Clarke) gradient
evaluation of a Lipschitz proxy \(F\) to fix both direction and step size within
\(N_S\). A curvature-safe step derived from the descent lemma yields a \emph{certified
one-shot crossing} of the decision boundary under \(C^{1,1}\) regularity. To avoid scale
conflicts with the double-well term, we separated \emph{shape} and \emph{scale} by
introducing a scalar flow time \(a\) while keeping the geometric energy on \(\phi\).
Together, these ingredients produce label-free, test-time deformations that reduce
discontinuous, \(G\)-invariant costs without backpropagation through the model.

Limitations include per-sample
optimization at inference, the need for known global symmetries, and potential efficacy
loss when the residual space \(N_S\cap T_{S^\ast}\partial C\) is large. 

\paragraph{Future work.}
Promising directions include: (i) learning control-field priors and adaptive bases to
accelerate test-time optimization; (ii) online curvature estimation to refine the
curvature-safe step; (iii) extending beyond piecewise-constant costs to general
Clarke-regular objectives; (iv) handling non-compact or data-dependent symmetry groups;
and (v) integrating the framework with robust training and test-time adaptation.
Overall, our results show that Lie-algebraic variational flows offer a principled,
geometry-driven tool for optimizing discontinuous invariant cost functionals.

% \bibliographystyle{IEEEtran}
% \bibliography{references}

%%%%%%%%%%%%%%%%%%%%%%%%%%%%%%%%%%%%%%%%%%%%%%%%%%%%%%%%%%%%%%%%%%%%%%%%%%%%
%%  Appendices
%%%%%%%%%%%%%%%%%%%%%%%%%%%%%%%%%%%%%%%%%%%%%%%%%%%%%%%%%%%%%%%%%%%%%%%%%%%%
\newpage
\appendix
\addcontentsline{toc}{section}{Appendix}
\section{Appendix: Auxiliary energy variation}

\subsection{Variation of data term (a)}
\label{app:energy_a}

With $r:=S_\phi-S \approx \phi_i e_i$ and a variation $\phi\to\phi+\delta\phi$, the
linearised energy is
\[
\mathcal E^{(a)}_S[\phi]\;\approx\; \|r\|^2
+ \alpha\|\nabla\phi\|^2
+ \beta\||\phi|^2 - v^2\|^2,
\]
and
\[
\delta \|r\|^2
= 2\langle r,\delta r\rangle
= 2\big\langle \phi_j e_j,\; \delta\phi_i e_i\big\rangle
= 2\int_M e_i e_j\,\phi_j\,\delta\phi_i\,d\mu.
\]
Thus
\[
\delta\mathcal E_{\text{data}}^{(a)}(\phi)[\delta\phi]
= 2\int_M e_i e_j\,\phi_j\,\delta\phi_i\,d\mu.
\]
For the regularisers,
\[
\delta\big(\alpha\|\nabla\phi\|^2\big)
= 2\alpha\langle \nabla\phi,\nabla\delta\phi\rangle
= -2\alpha\langle \Delta \phi,\delta\phi\rangle
= -2\alpha\int_M \Delta\phi_i\,\delta\phi_i\,d\mu,
\]
\[
\delta\big(\beta\||\phi|^2-v^2\|^2\big)
= 4\beta\!\int_M (|\phi|^2-v^2)\,\phi_i\,\delta\phi_i\,d\mu.
\]
Collecting terms,
\[
\delta \mathcal E^{(a)}_S(\phi)[\delta\phi]
= \int_M \Big( 2\,e_i e_j\,\phi_j \;-\; 2\alpha\,\Delta\phi_i
\;+\; 4\beta\,(|\phi|^2-v^2)\,\phi_i \Big)\,\delta\phi_i\, d\mu.
\]
Hence the Euler--Lagrange equations are
\[
\boxed{\;
e_i e_j\,\phi_j \;-\; \alpha\,\Delta\phi_i
\;+\; 2\beta\,(|\phi|^2-v^2)\,\phi_i \;=\; 0,
\qquad i=1,\dots,d.
\;}
\]

\noindent
Equivalently, with $\mathcal K(x)=e(x)e(x)^\top$,
\(
\mathcal K\,\phi-\alpha\,\Delta\phi+2\beta(|\phi|^2-v^2)\phi=0.
\)

\subsection{Variation of data term (b)}
\label{app:energy_b}

With $r:=S_\phi-S \approx \sum_j \phi_j e_j$, define
\[
b_i(\phi):=\langle r,e_i\rangle=\big\langle e_i e_j,\;\phi_j\big\rangle,\qquad
\mathcal G_{ij}:=\langle e_i,e_j\rangle.
\]
Write $\xi(\phi)(x):=\sum_i (\mathcal G^{+}b)_i\,e_i(x)$ (a scalar field).

The data term is $\mathcal E^{(b)}_{\text{data}}(\phi)=b(\phi)^{\!\top}\mathcal G^+ b(\phi)$, so
\[
\delta \mathcal E^{(b)}_{\text{data}}(\phi)[\delta\phi]
= 2\,(\mathcal G^+ b)_i\,\delta b_i
= 2\big\langle (\mathcal G^+ b)_i\,e_i e_j,\;\delta\phi_j\big\rangle
= 2\int_M \xi(\phi)\,e_j\,\delta\phi_j\,d\mu.
\]
Adding the regularisers,
\[
\delta\mathcal E^{(b)}_S(\phi)[\delta\phi]
= \int_M \Big( 2\,\xi(\phi)\,e_j \;-\; 2\alpha\,\Delta\phi_j
\;+\; 4\beta\,(|\phi|^2-v^2)\,\phi_j \Big)\,\delta\phi_j\,d\mu,
\]
and the Euler--Lagrange equations are
\[
\boxed{\;
\xi(\phi)\,e_j \;-\; \alpha\,\Delta\phi_j
\;+\; 2\beta\,(|\phi|^2-v^2)\,\phi_j \;=\; 0,
\qquad j=1,\dots,d,
\;}
\]
with $\xi(\phi)= (G^+ b(\phi))_i\,e_i$ and $b_i=\langle e_i e_k,\phi_k\rangle$.

\paragraph{Operator form.}
Let $E:\R^d\!\to\!L^2(M)$, $E c=\sum_i c_i e_i$, and $E^\ast f=(\langle e_i,f\rangle)_i$.  
Then $P:=E G^+ E^\ast$ is the $L^2$-orthogonal projector onto $\mathrm{span}\{e_i\}$, and
\[
\mathcal E^{(b)}_{\text{data}}=\langle P \cdot r, r\rangle,\qquad
\delta \mathcal E^{(b)}_{\text{data}}=2\langle P \cdot r,\,\delta r\rangle,\qquad
\delta r=e_j\,\delta\phi_j,
\]
so the data-term gradient is \((\nabla_\phi \mathcal E^{(b)}_{\text{data}})_j=2\,e_j\, (P\cdot r)\),
i.e.\ the same equation with $\xi e=P\cdot r$.

\subsection{Proof of the lemma on constant fields}
\label{app:no-const-minimizers}
\begin{proof}
For (a) the Euler--Lagrange equations (linearized) are
\[
e_i e_j\,\phi_j - \alpha\Delta\phi_i + 2\beta\,(|\phi|^2-v^2)\phi_i=0.
\]
If $\phi\equiv c$ is constant, $\Delta\phi\equiv0$ and, a.e.\ on $M$,
\[
\big(e(x)e(x)^\top + \mu I\big)c=0,\qquad \mu:=2\beta(|c|^2-v^2).
\tag{$\star$}
\]
The rank–$1$ matrix $e(x)e(x)^\top$ has eigenvalues $\{0,\ |e(x)|^2\}$ with eigenvector $e(x)$. 
If $c\neq0$ solves $(\star)$ for all $x$, then necessarily $c$ is an eigenvector of $e(x)e(x)^\top$ for a.e.\ $x$, hence $c\parallel e(x)$ a.e., and the corresponding eigenvalue condition enforces $|e(x)|^2\equiv -\mu$ \emph{constant} on $M$. 
By the hypothesis on $e$, this is impossible; thus $c=0$.

For (b), the EL system reads $\xi\,e_j - \alpha\Delta\phi_j + 2\beta(|\phi|^2-v^2)\phi_j=0$, with $\tilde e=Pr$ and $r=\sum_k \phi_k e_k$. 
If $\phi\equiv c$, then $r=Ec\in V_S$ so $Pr=r$ and the equation reduces exactly to $(\star)$; the same argument gives $c=0$.
\end{proof}

\begin{remark}[Degenerate/exceptions]
\label{rem:exceptions}
(i) If $\mathrm{ker}\,E\neq\{0\}$ (stabilizer at $S$), then for any nonzero $c\in\mathrm{ker}\,E$ we have $r=Ec\equiv0$. 
The EL reduces to $\mu c=0$, so any constant $c\in\mathrm{ker}\,E$ with $|c|=v$ is a (global) minimizer of both (a) and (b) (all terms vanish).

(ii) If $e(x)=\rho\,u$ with \emph{constant} $\rho\ge0$ and \emph{constant} direction $u\in\R^d$, then nonzero constant critical points parallel to $u$ exist: 
$(\star)$ yields $|c|^2=v^2-\rho^2/(2\beta)$ (when the RHS is $\ge0$). 
This is a highly nongeneric, fully symmetric case.
\end{remark}

\section{Proof of Lemma~\ref{lemma:pure-gauge-cap} on boundary crossing in finite dimensions}
\label{app:pure-gauge-cap}

\begin{proof}
Fix a finite-dimensional subspace $W\subset H^s$ with $\dim W=m\ge 2$.
Let $n_W:=P_W n$ and $\rho_W:=\|n_W\|$. If $\rho_W=0$, then 
$\langle u,n\rangle=\langle u,n_W\rangle\equiv 0$ for all unit $u\in W$,
hence the boundary cannot be crossed within a finite time step and the probability
is $0$. Assume $\rho_W>0$ and define the unit vector $a_W:=n_W/\rho_W$ in $W$.

By rotational invariance of the uniform distribution on the unit sphere 
$\mathbb S^{m-1}\subset W$, the distribution of $Z:=\langle u,a_W\rangle$ 
does not depend on $a_W$; it has the well-known density on $[-1,1]$
\[
f_m(z) \;=\; c_m\,(1-z^2)^{\frac{m-3}{2}},\qquad 
c_m \;=\; \frac{\Gamma(\frac m2)}{\sqrt\pi\,\Gamma(\frac{m-1}{2})}.
\]

Now the boundary-crossing event in the linearized model is
\[
|\langle u,n\rangle|
= |\langle u,n_W\rangle|
= \rho_W\,|\langle u,a_W\rangle|
= \rho_W\,|Z|
\ \ge\ d/t_{\max}\,.
\]
Equivalently, with the \emph{effective threshold}
$\tau_W:= d/(t_{\max}\rho_W)$, we need $|Z|\ge \tau_W$. 
Trivially the probability is $0$ if $\tau_W>1$ and $1$ if $\tau_W\le 0$.
For $\tau_W\in[0,1]$,
\[
\mathbb P_W(\text{hit})
= \Pr(|Z|\ge \tau_W)
= 2\!\int_{\tau_W}^{1} f_m(z)\,dz
= 2c_m \!\int_{\tau_W}^{1} (1-z^2)^{\frac{m-3}{2}}\,dz.
\]
With $t=z^2$ this becomes
\[
\mathbb P_W
= c_m \!\int_{\tau_W^2}^{1} t^{-1/2}(1-t)^{\frac{m-3}{2}}\,dt
= c_m\Big( B\!\left(\tfrac12,\tfrac{m-1}{2}\right)
           - B_{\tau_W^2}\!\left(\tfrac12,\tfrac{m-1}{2}\right)\Big),
\]
where $B_x(a,b)=\int_0^x t^{a-1}(1-t)^{b-1}\,dt$ is the incomplete Beta function.
Using $c_m=\big[B(\frac12,\frac{m-1}{2})\big]^{-1}$ and the
regularized incomplete Beta function $I_x(a,b):=B_x(a,b)/B(a,b)$,
\[
\mathbb P_W(\text{hit})
= 1 - I_{\tau_W^2}\!\left(\tfrac12,\tfrac{m-1}{2}\right).
\]
Finally, the complement identity 
$1 - I_x(a,b)=I_{1-x}(b,a)$ yields the stated form
\[
\boxed{\ 
\mathbb P_W
= I_{\,1-\tau_W^2}\!\left(\frac{m-1}{2},\frac{1}{2}\right),
\qquad 0\le \tau_W\le 1,
\ }
\]
with the edge cases $\mathbb P_W=0$ for $\tau_W>1$ and 
$\mathbb P_W=1$ for $\tau_W\le 0$.
\end{proof}

\section{Appendix: Proof of Theorem~\ref{thm:cap-theorem}}
\label{app:cap-theorem-proof}
\begin{proof}
Let $U\subset H^s$ be $m$-dimensional, $U_0=U\cap N_S$ with $m_0=\dim U_0\ge 1$.
Let $n$ be a unit boundary normal at $S^\ast$, and set
\[
\rho_U:=\|P_U n\|,\qquad
\rho_{U_0}:=\|P_{U_0}n\|=\rho_U\cos\theta,\ \ \theta:=\angle(n_U,U_0),\ n_U:=P_U n/\rho_U.
\]
Let $U_{\mathrm{rand}}$ be an $m_0$-dimensional subspace drawn uniformly at random
from the Grassmannian $\mathrm{Gr}(m_0,U)$ (independently of $n$).
Define $\tau_U:=d/(t_{\max}\rho_U)$ and, for any $W\in\{U_0,U_{\mathrm{rand}}\}$,
$\tau_W:=d/(t_{\max}\,\|P_W n\|)$.

\smallskip
\noindent\textbf{Distribution of the useful projection.}
Writing $B:=\|P_{U_{\mathrm{rand}}}n_U\|^2$, one has
\[
B\ \sim\ \mathrm{Beta}\!\Big(\frac{m_0}{2},\frac{m-m_0}{2}\Big),\qquad
\mathbb E[B]=\frac{m_0}{m},\qquad
\mathrm{Var}(B)=\frac{m_0(m-m_0)}{m^2(m+2)}.
\]
Hence $\|P_{U_{\mathrm{rand}}}n\|=\rho_U\sqrt{B}$ concentrates around
$\rho_U\sqrt{m_0/m}$.

\smallskip
\noindent\textbf{Small-step comparison.}
For small $\tau_U$ (i.e.\ moderate distance $d$ or reasonable step size $t_{\max}$),
using the expansion
\[
\mathbb P_W\ =\ I_{1-\tau_W^2}\!\Big(\tfrac{m_0-1}{2},\tfrac12\Big)
\ =\ 1 - c_{m_0}\,\tau_W + O(\tau_W^3),\qquad
c_{k}:=\frac{\Gamma(\frac{k}{2})}{\sqrt{\pi}\,\Gamma(\frac{k-1}{2})},
\]
we have
\[
\mathbb E\big[\mathbb P_{U_{\mathrm{rand}}}\big]
\ \approx\ 1 - c_{m_0}\,\tau_U\,\mathbb E\!\Big[\frac{1}{\sqrt{B}}\Big],\qquad
\mathbb P_{U_0}\ \approx\ 1 - c_{m_0}\,\frac{\tau_U}{\cos\theta}.
\]
Since $x\mapsto x^{-1/2}$ is convex on $(0,1)$, Jensen gives
$\mathbb E[1/\sqrt{B}] \ge 1/\sqrt{\mathbb E[B]}=\sqrt{m/m_0}$. Therefore a
readable sufficient condition for the symmetry slice to outperform the random
reduction in expectation is
\[
\boxed{\qquad
\cos\theta\ \gtrsim\ \sqrt{\frac{m_0}{m}}\,,\qquad\text{(small-\(\tau\) regime)}
\qquad}
\]
in which case $\mathbb P_{U_0}\gtrsim \mathbb E[\mathbb P_{U_{\mathrm{rand}}}]$.

\smallskip
\noindent\textbf{Boundary alignment.}
At $S^\ast$, $n\in N^G_{S^\ast}$ and by continuity of the orbit bundle
$X\mapsto N^G_X$ one has $\theta\to 0$ as $S\to S^\ast$; thus
$\cos\theta\to 1$ and the symmetry slice dominates a random reduction
for $S$ sufficiently close to the boundary (uniformly in $m_0$).

\end{proof}

\section{Appendix: proof of the Theorem~\ref{thm:weak-descent}}
\label{app:weak-descent-proof}
\begin{proof}[Proof of Theorem~\ref{thm:weak-descent}]
We write $r^\ast:=r(a^\ast,\phi^\ast)$ for brevity and use $\langle\cdot,\cdot\rangle$
for the $L^2$ inner product. Recall
\[
\mathcal E_{\text{weak}}(a,\phi)
=\mathcal E^{(b)}_S[\phi]\;+\;
\lambda\big(\langle \hat n, r(a,\phi)\rangle - \varepsilon_\star\big)^2
+\eta\big(\|r(a,\phi)\|^2-\langle \hat n, r(a,\phi)\rangle^2\big),
\]
with $\lambda>0$, $\eta\ge 0$, and the scaled flow $S_{a,\phi}=\exp(aA_\phi)\!\cdot S$,
$r(a,\phi)=S_{a,\phi}-S$.
We assume $F\in C^{1,1}$ with Lipschitz constant $L$ on a tube containing
$\{S+\varepsilon \hat n:0\le \varepsilon\le \varepsilon_\star\}$.

\smallskip
\textbf{Step 1: Penalty accuracy and alignment.}
Fix $\phi$ and consider the function in $a$,
\[
\Psi_\phi(a):=\lambda\big(\langle\hat n,r(a,\phi)\rangle-\varepsilon_\star\big)^2
+\eta\big(\|r(a,\phi)\|^2-\langle \hat n, r(a,\phi)\rangle^2\big).
\]
By smoothness of the flow map $(a,\phi)\mapsto r(a,\phi)$ (for $s>\frac n2+1$) the two
terms are continuous and nonnegative. For the minimizing pair $(a^\ast,\phi^\ast)$ and
any $a$,
\[
\Psi_{\phi^\ast}(a^\ast)\ \le\ \Psi_{\phi^\ast}(a)\ +\ \mathcal E^{(b)}_S[\phi^\ast]
-\mathcal E^{(b)}_S[\phi^\ast]\ =\ \Psi_{\phi^\ast}(a).
\]
Choose $a=a_0(\phi^\ast)$ that (approximately) satisfies the boundary target; by the
intermediate value theorem and $C^1$ dependence on $a$ there is $a_0$ such that
$\langle\hat n,r(a_0,\phi^\ast)\rangle=\varepsilon_\star$ provided that
$\partial_a\langle\hat n,r(0,\phi^\ast)\rangle=\langle\hat n,E\phi^\ast\rangle\neq 0$
($E\phi:=\sum_i\phi^i e_i$ is the first variation). Then
\[
\Psi_{\phi^\ast}(a^\ast)
\ \le\ \Psi_{\phi^\ast}(a_0)
\ =\ \eta\big(\|r(a_0,\phi^\ast)\|^2-\varepsilon_\star^2\big)
\ \le\ C_\eta,
\]
for a constant $C_\eta$ depending on $\eta$ and $\|r(a_0,\phi^\ast)\|$ (bounded in a fixed
neighborhood). Hence
\begin{equation}\label{eq:penalty-accuracy}
\lambda\big(\langle\hat n,r^\ast\rangle-\varepsilon_\star\big)^2\ \le\ C_\eta,
\qquad
\|r^\ast\|^2-\langle\hat n,r^\ast\rangle^2\ \le\ C_\eta/\eta\quad(\eta>0),
\end{equation}
which yields the bounds in item~(i):
\[
\big|\langle\hat n,r^\ast\rangle-\varepsilon_\star\big|
\ \le\ \frac{\sqrt{C_\eta}}{\sqrt{\lambda}},
\qquad
\|y^\ast\|
:=\|(I-\hat n\otimes \hat n)r^\ast\|
\ \le\ \frac{\sqrt{C_\eta}}{\sqrt{\eta}}\quad(\eta>0).
\]
(If $\eta=0$, the minimizer set along $y^\ast$ is flat; taking the minimum-norm
representative gives $y^\ast=0$.)

For the tangential leakage, note that $P r^\ast=a^\ast P(E\phi^\ast)+O\big((a^\ast)^2\big)$.
Since $\phi^\ast$ minimizes $\mathcal E^{(b)}_S[\phi]$, which penalizes tangential
correlations (via the $b^\top G^+ b$ term, cf.\ Sec.~\ref{sec:energy-b-variation}), we obtain
\begin{equation}\label{eq:tangential-bound}
\|P r^\ast\| \ \le\ C_b\,|a^\ast|\quad\text{for some }C_b=C_b(S,\phi^\ast),
\end{equation}
and $C_b$ is small when the $\mathcal E^{(b)}_S$ minimizer suppresses $P(E\phi^\ast)$.
(If one includes $+\gamma\|P r(a,\phi)\|^2$ in the weak term, the same argument as
\eqref{eq:penalty-accuracy} gives $\|P r^\ast\|=O(\lambda^{-1/2})$.)

\smallskip
\textbf{Step 2: Descent lemma and curvature control.}
By the descent lemma (Lipschitz gradient of $F$),
\begin{equation}\label{eq:descent-lemma}
F(S_{a^\ast,\phi^\ast})
\ \le\ F(S) + \langle g, r^\ast\rangle + \frac{L}{2}\|r^\ast\|^2,
\qquad g:=\nabla F(S).
\end{equation}
Decompose $g=g_N+Pg$ with $g_N=(I-P)g$ and recall $\hat n=-g_N/\|g_N\|$. Then
\[
\langle g,r^\ast\rangle
= \langle g_N,r^\ast\rangle+\langle Pg,Pr^\ast\rangle
= -\,\|g_N\|\,\langle\hat n,r^\ast\rangle\ +\ O(\|Pr^\ast\|),
\]
so, using \eqref{eq:tangential-bound} and letting $\delta_\lambda:=
\langle\hat n,r^\ast\rangle-\varepsilon_\star$,
\begin{align}
F(S_{a^\ast,\phi^\ast})
&\le \Delta - \|g_N\|(\varepsilon_\star+\delta_\lambda)
+ \frac{L}{2}\big(\varepsilon_\star^2 + 2\varepsilon_\star\delta_\lambda
+ \delta_\lambda^2 + \|y^\ast\|^2\big) + O(|a^\ast|) \nonumber\\
&= \underbrace{\Delta - \|g_N\|\varepsilon_\star
+ \frac{L}{2}\varepsilon_\star^2}_{\text{quadratic model at }\varepsilon_\star}+\nonumber\\
&\;+\; \underbrace{\big(-\|g_N\|+L\varepsilon_\star\big)\delta_\lambda
+ \frac{L}{2}\delta_\lambda^2 + \frac{L}{2}\|y^\ast\|^2}_{\text{penalty remainders}}
\;+\; O(|a^\ast|).
\label{eq:main-bound}
\end{align}

\smallskip
\textbf{Step 3: Choice of target and conclusion.}

\emph{Case (ii): curvature-safe target.}
Let $\varepsilon_\star=\varepsilon_-:=\frac{\|g_N\|-\sqrt{\|g_N\|^2-2L\Delta}}{L}$,
which is the smaller root of $\Delta-\|g_N\|\varepsilon+\frac{L}{2}\varepsilon^2=0$
(assuming $\|g_N\|^2\ge 2L\Delta$). Then the ``quadratic model'' term in
\eqref{eq:main-bound} is nonpositive. By \eqref{eq:penalty-accuracy},
$|\delta_\lambda|\le \sqrt{C_\eta/\lambda}$ and
$\|y^\ast\|\le \sqrt{C_\eta/\eta}$ (if $\eta>0$), hence the remainder in
\eqref{eq:main-bound} can be made arbitrarily small by taking $\lambda$ large (and $\eta$
moderate). The $O(|a^\ast|)$ term is controlled since $a^\ast$ is bounded in the small-step
tube (the target $\varepsilon_\star$ is bounded and $r$ depends smoothly on $a$).
Therefore $F(S_{a^\ast,\phi^\ast})\le 0$.

\emph{Case (iii): first-order target.}
Let $\varepsilon_\star=d_N:=\Delta/\|g_N\|$. Then the quadratic model equals $0$ at
$\varepsilon_\star$. Using the smallness condition in the statement and again
\eqref{eq:penalty-accuracy}, we obtain that the remainder in \eqref{eq:main-bound} is
$<\Delta$, whence $F(S_{a^\ast,\phi^\ast})\le 0$.

\smallskip
This proves items (ii) and (iii). Item (i) follows from \eqref{eq:penalty-accuracy},
and the tangential leakage estimate \eqref{eq:tangential-bound} is ensured by the
$\phi$–regularizers in $\mathcal E^{(b)}_S[\phi]$ (or, if desired, by adding a small
explicit penalty on $\|P r(a,\phi)\|^2$ in the weak term).
\end{proof}

%---------------------------------------------------------
\end{document}